\documentclass[letterpaper]{article} 
\usepackage[preprint]{aaai2027}  
\usepackage[hyphens]{url}  
\usepackage{graphicx} 
\usepackage{natbib}  
\usepackage{caption} 
\usepackage{algorithm}
\usepackage{algorithmic}
\usepackage{amsmath}
\usepackage{amssymb}
\usepackage{newfloat}
\usepackage{listings}
\DeclareCaptionStyle{ruled}{labelfont=normalfont,labelsep=colon,strut=off} 
\floatstyle{ruled}
\newfloat{listing}{tb}{lst}{}
\floatname{listing}{Listing}

\usepackage{booktabs}
\usepackage{multirow}
\title{Disentangling 3D Modeling from Spatial Reasoning}
\author{
    Haoze Sun\textsuperscript{\rm 1}, Jiequan Cui\textsuperscript{\rm 1}\corresponding, Qingshan Xu\textsuperscript{\rm 2}, Richang Hong\textsuperscript{\rm 1}
}
\affiliations{
    \textsuperscript{\rm 1} HFUT, \textsuperscript{\rm 2} USTC

}

\begin{document}

\maketitle

\begin{abstract}
In this work, we explore an alternative paradigm for spatial reasoning by explicitly disentangling 3D perception from reasoning, rather than jointly acquiring implicit 3D perception and reasoning through large-scale training. Our key observation is that modern perception models excel at estimating continuous 3D geometry, whereas large language models (LLMs) are particularly effective at compositional and symbolic reasoning. Motivated by these complementary strengths, we propose the \textit{Disentangled Spatial Reasoner (DiSR)}, a simple yet effective framework that reconstructs the physical world into structured 3D evidence using off-the-shelf expert perception models and fine-tunes an LLM with LoRA to perform reasoning solely over this explicit geometric evidence. Without large-scale 3D VQA training or complex tool-use policies, \textit{DiSR} achieves competitive performance on popular spatial reasoning benchmarks. Beyond its strong performance, \textit{DiSR} offers improved interpretability, modularity, and computational efficiency, 
demonstrating that explicit separation of perception and reasoning is a scalable and effective alternative paradigm to end-to-end modeling for spatial intelligence.
\end{abstract}


\section{Introduction}
\label{sec:intro}

Spatial reasoning is a fundamental capability for intelligent systems to understand, interpret, and interact with the physical 3D world. Unlike conventional visual recognition, which focuses on identifying objects and their semantic attributes, spatial reasoning requires models to infer geometric properties, spatial relationships, and physical constraints from visual observations. Such reasoning is essential in a wide range of applications, including embodied agents, robotic manipulation, autonomous driving, augmented reality, and human-robot interaction, where decision-making depends not only on \textit{what} objects are present but also on \textit{where} they are and \textit{how} they relate to one another in 3D space. As multimodal large language models (MLLMs) continue to demonstrate strong progress in visual understanding and language reasoning~\citep{liu2023visual,achiam2023gpt,bai2025qwen3}, enabling reliable spatial reasoning has become a key step toward building general-purpose AI systems.

\begin{figure*}[t]
    \centering
    \includegraphics[width=0.90\textwidth]
    {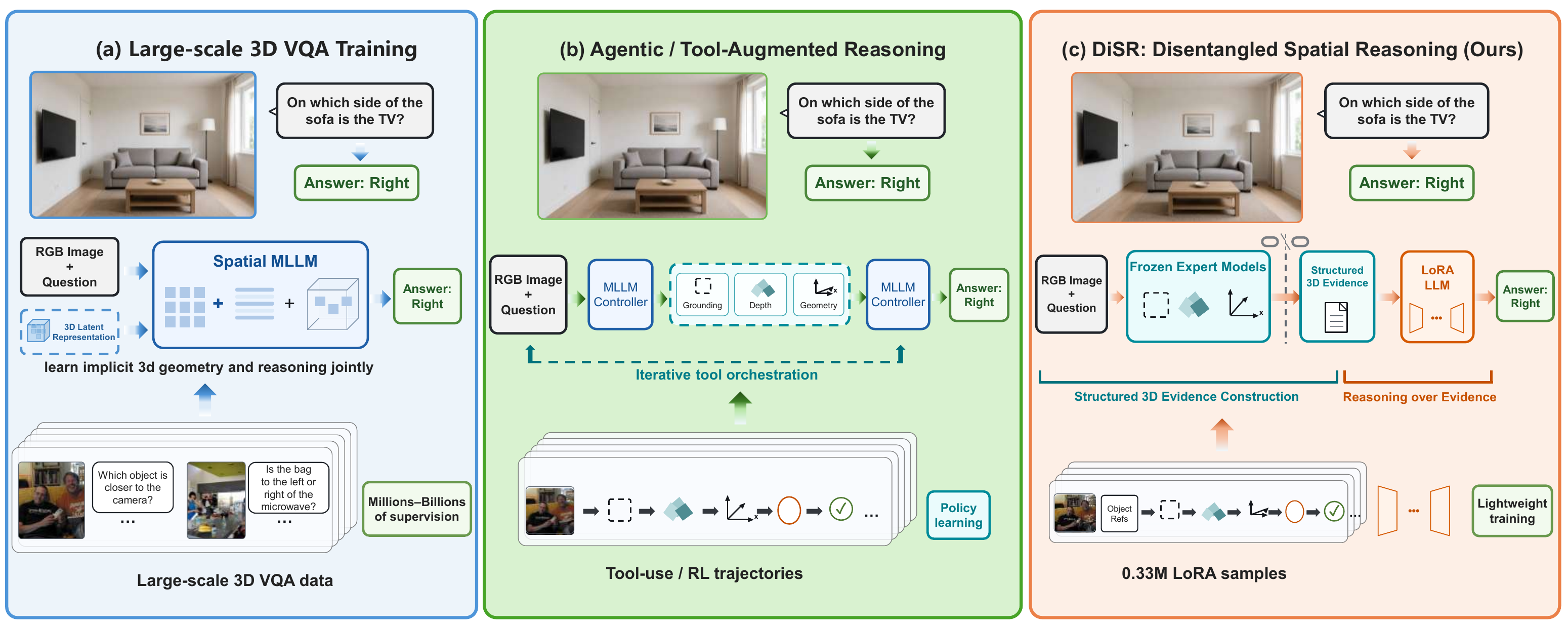}
    \caption{
    Comparison of spatial reasoning paradigms. (a) Methods jointly acquire implicit geometric perception and spatial reasoning through extensive spatial supervision,
    whereas (b) tool-augmented agents learn tool-use policy and invoke external tools through iterative execution.
    (c) \textit{DiSR} instead constructs explicit 3D evidence and then performs reasoning with an LLM solely on the structured evidence,
    thereby disentangling 3D modeling from spatial reasoning.
    }
    \label{fig:paradigm_comparison}
    \vspace{-0.1in}
\end{figure*}

Despite these advances, achieving robust spatial intelligence with MLLMs remains challenging. Existing approaches either train MLLMs on large-scale 3D visual question answering (3D VQA) datasets to acquire geometric understanding~\citep{chen2024spatialvlm,cheng2024spatialrgpt,ma2025spatialllm,ma2026spatialreasoner,liang2026hispatial}, or align explicit 3D latent representations with MLLMs through multimodal learning~\citep{hong20233d,zhu20233d,xu2024pointllm,huang2023embodied} as shown in Figure~\ref{fig:paradigm_comparison}(a). Despite differences, these approaches share a common formulation: jointly optimizing implicit geometric perception and spatial reasoning. Consequently, 3D knowledge is embedded in latent representations rather than explicitly reconstructed into interpretable geometric structures. Moreover, such large-scale training often requires substantial computational resources, limiting the scalability and efficiency of spatial reasoning systems. Recently, agentic frameworks shown in Figure~\ref{fig:paradigm_comparison}(b) have explored the use of external perception tools to mitigate these challenges, but introduce additional complexity by requiring MLLMs to learn effective tool-use strategies through reinforcement learning or strong foundation models~\citep{ma2024spatialpin,chen2026spacetools,han2025tiger,chen2026geometrically}.

\noindent{\bf Disentangling 3D Perception from Spatial Reasoning.}
Rather than further coupling perception and reasoning, we explore an alternative paradigm based on explicit disentanglement. This idea is motivated by the complementary strengths of modern perception models and large language models (LLMs): specialized perception models have demonstrated strong capabilities in estimating continuous geometric properties, such as depth, object poses, and 3D layouts, while LLMs excel at compositional and symbolic reasoning over structured information.  
These observations raise a natural question: can spatial intelligence be achieved by allowing each component to specialize in its respective capability, rather than requiring a single MLLM to jointly learn both?

Based on this insight, we propose to explicitly separate 3D perception from spatial reasoning as shown in Figure~\ref{fig:paradigm_comparison}(c). Specifically, we leverage off-the-shelf expert perception models to reconstruct the physical world into structured 3D evidence, including 3D locations, object orientations, and object sizes. An LLM with LoRA is then fine-tuned to reason solely over this explicit geometric evidence. By introducing an interpretable intermediate textual evidence between perception and reasoning, our method 
offers a simple, efficient, and modular alternative to end-to-end joint modeling.

\noindent{\bf Disentangled Spatial Reasoner (DiSR).}
To instantiate this idea, we propose the \textit{Disentangled Spatial Reasoner (DiSR)}, illustrated in Figure~\ref{fig:paradigm_comparison}(c). \textit{DiSR} decomposes spatial reasoning into two steps: (1) \textit{structured 3D evidence construction}, where specialized perception models produce structured geometric evidence, including object 3D locations, object sizes, and object orientations; and (2) \textit{reasoning over structured 3D evidence}, where an LLM performs structured reasoning solely over the extracted geometric evidence inputs without images to produce the final answer.

This disentangled design yields several practical advantages. First, by offloading geometric perception to specialized perception models, \textit{DiSR} significantly reduces the need for large-scale 3D supervision 
during MLLM training, leading to more efficient training and lower computational cost, as summarized in Table~\ref{tab:3dsrbench_main}. Second, the modular formulation improves interpretability and diagnosability: since structured 3D evidence is explicitly available, reasoning errors can be clearly separated from perception errors, as demonstrated by the controlled diagnostic analysis in Table~\ref{tab:cvbench3d_analysis}. Third, the framework is highly extensible, allowing improvements in either perception models or LLMs to be incorporated independently without retraining the full system, as evidenced by the cross-backbone results in Table~\ref{tab:extensibility_disr}.

\noindent{\bf Experimental Results.}
We evaluate \textit{DiSR} on three representative benchmarks, including 3DSRBench~\citep{ma20253dsrbench}, SPAR-Bench~\citep{zhang2026flatland}, and CV-Bench-3D~\citep{tong2024cambrian}. Despite using significantly less computational cost and fewer trainable parameters, \textit{DiSR} consistently outperforms prior methods that rely on large-scale 3D training, such as SpatialRGPT~\citep{cheng2024spatialrgpt} and SpatialReasoner~\citep{ma2026spatialreasoner}. Specifically, we achieve new state-of-the-art performance on 3DSRBench and SPAR-Bench, outperforming previous best methods by \textbf{3.77\%} and \textbf{1.70\%}, respectively.
These results demonstrate that explicitly disentangling 3D perception from reasoning is an effective and data-efficient paradigm for spatial intelligence.

Overall, our contributions are summarized as follows:
\begin{itemize}
    \item We propose to disentangle 3D perception from spatial reasoning and introduce \textit{DiSR}, which leverages 3D perception expert models to construct explicit evidence and reason solely over the structured evidence with LLMs.
    \item We demonstrate that \textit{DiSR} with this disentangled design achieves strong performance while using significantly less training data and computational resources.
    \item We validate \textit{DiSR} on popular spatial reasoning benchmarks, where it consistently outperforms prior large-scale training approaches, achieving new state-of-the-art results on the challenging 3DSRBench and SPAR-Bench.
\end{itemize}
\section{Related Work}

\noindent{\bf Spatial VLMs with Large-scale 3D Supervision.}
A dominant approach to spatial reasoning is to enhance MLLMs through large-scale 3D supervision. Early works construct 3D visual question answering (3D VQA) datasets from monocular images and fine-tune VLMs to implicitly acquire geometric knowledge. For example, SpatialVLM~\citep{chen2024spatialvlm} introduces Internet-scale 3D VQA data, while SpatialRGPT~\citep{cheng2024spatialrgpt}  
adopts region-level 3D scene graphs and depth 
cues to improve geometric representation learning. Subsequent studies, including SpatialLLM~\citep{ma2025spatialllm}, SpatialReasoner~\citep{ma2026spatialreasoner}, and HiSpatial~\citep{liang2026hispatial}, explore improved 3D-aware data, training strategies, reinforcement learning, and RGB-D representations to further enhance spatial reasoning capabilities.

Despite their differences, these methods share a common formulation: jointly learning implicit 3D perception and spatial reasoning within a single MLLM through large-scale optimization. This design often requires substantial 3D supervision and computational resources, while coupling perception and reasoning into an implicit representation that makes error diagnosis challenging. In contrast, \textit{DiSR} adopts a disentangled paradigm by delegating geometric perception to specialized expert models and training the LLM solely to reason over explicit, structured 3D evidence. This design enables efficient learning while preserving the complementary strengths of perception models and language models.

\noindent{\bf Aligning MLLMs with 3D Representations.}
Another line of research equips MLLMs with spatial intelligence by aligning them with explicit 3D representations. Instead of relying solely on 3D VQA supervision, these methods bridge the modality gap between geometric representations and language models through dedicated encoders, alignment objectives, or instruction tuning. For example, 3D-LLM~\citep{hong20233d} projects 3D scene features into the language space to support tasks such as question answering, grounding, navigation, and dialogue. 3D-VisTA~\citep{zhu20233d} pre-trains unified 3D vision-language representations over RGB-D scenes and scene-text pairs for a variety of downstream tasks. PointLLM~\citep{xu2024pointllm} aligns point-cloud encoders with LLMs using point-text instruction tuning, enabling language models to understand object-level 3D geometry. LEO~\citep{huang2023embodied} further extends this paradigm to embodied agents through joint 3D vision-language alignment and vision-language-action instruction tuning.
These methods aim to bridge the modality gap between 3D representations and MLLMs, which suffer from similar problems with methods training on large-scale 3D supervision.

\noindent{\bf Agentic Spatial Reasoning.}
Another line of research enhances spatial reasoning by augmenting MLLMs with external perception models, geometric tools, or formal reasoning modules. Instead of encoding all geometric knowledge within the language model, these approaches enable MLLMs to acquire spatial capabilities through tool invocation. For example, SpatialPIN~\citep{ma2024spatialpin} improves spatial reasoning in a training-free manner by leveraging priors from multiple 3D foundation models. SpaceTools~\citep{chen2026spacetools} trains MLLMs to coordinate multiple perception and robotics tools through reinforcement learning, while TIGeR~\citep{han2025tiger} equips MLLMs with geometric computation tools for precise spatial operations. GCA~\citep{chen2026geometrically} further incorporates formal constraints to connect semantic understanding with geometric reasoning.

These methods demonstrate the effectiveness of external geometric computation for spatial reasoning. However, they require learning tool-use policies, orchestrating multi-step interactions, or designing task-specific constraints. In contrast, \textit{DiSR} adopts a fixed and interpretable interface between perception and reasoning: expert perception models reconstruct the physical world into structured 3D evidence, and the LLM reasons directly over this explicit evidence. This design avoids the complexity of tool orchestration while enabling simpler reasoning, lightweight training, and clearer attribution of errors to either perception or reasoning.

\label{sec:related_work}
\section{Method}
\label{sec:method}

\begin{figure*}[t]
    \centering
    \includegraphics[
        width=0.9\linewidth,
        trim=0 4mm 0 0,
        clip
    ]{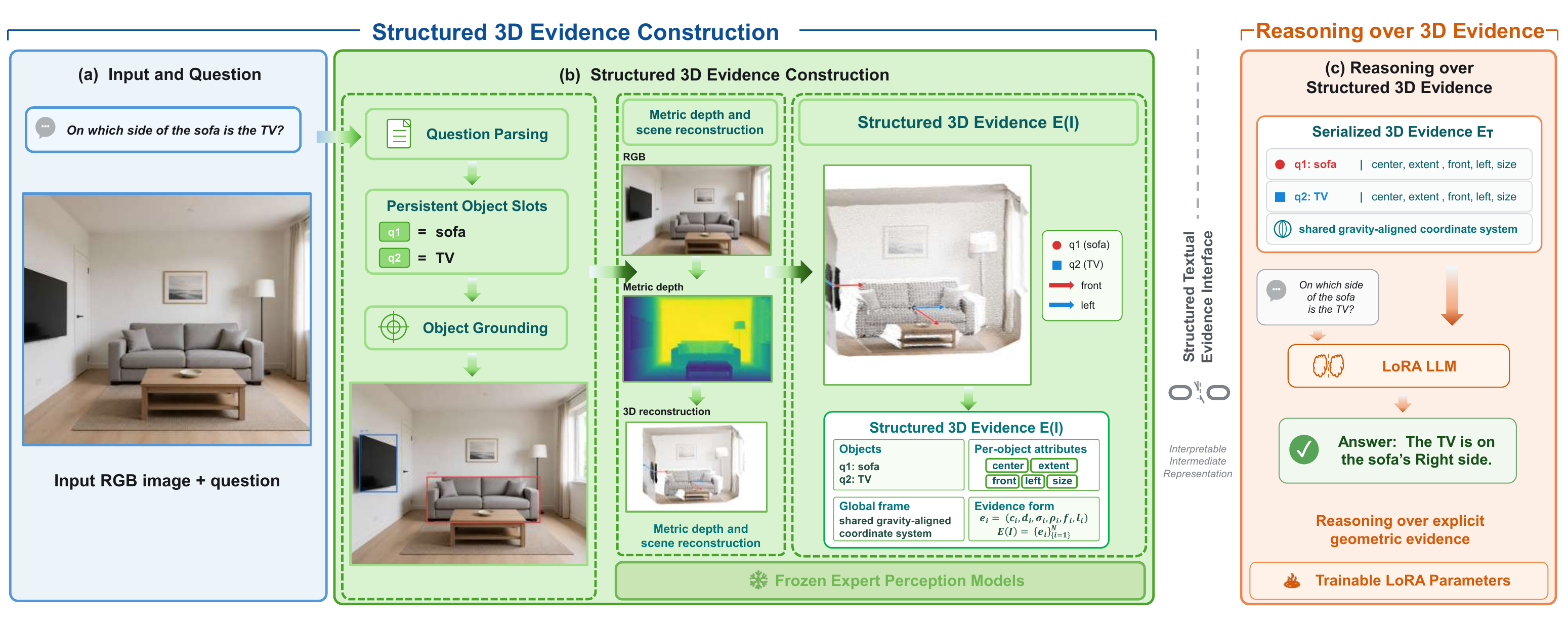}
    \vspace{-0.06in}
    \caption{Overview of DiSR.}
    \label{fig:disr_overview}
    \vspace{-0.1in}
\end{figure*}

Modern perception models have demonstrated remarkable capabilities in estimating continuous geometric properties, whereas LLMs excel at compositional and symbolic reasoning. Motivated by these complementary strengths, we propose the \textit{Disentangled Spatial Reasoner (DiSR)}, which formulates spatial reasoning as a two-stage process: reconstructing the physical world into structured 3D evidence, followed by reasoning over this explicit geometric evidence. Rather than jointly optimizing perception and reasoning, \textit{DiSR} explicitly separates the two through an interpretable intermediate textual evidence, resulting in a simple, modular, and efficient framework. We first present an overview of DiSR and then describe each component in detail.

\subsection{Overview of DiSR}
\label{sec:method_overview}
As illustrated in Figure~\ref{fig:disr_overview}, the key idea
of \textit{DiSR} is to explicitly separate 3D perception from spatial
reasoning through structured intermediate 3D evidence.
Instead of requiring a single MLLM to jointly learn geometric
perception and reasoning from large-scale 3D VQA data, \textit{DiSR}
assigns each capability to a specialized component: expert
perception models are responsible for constructing structured
3D evidence, while an LLM focuses on reasoning solely over
the resulting geometric representation.

Given an input image $I$ and a spatial reasoning question $Q$, \textit{DiSR} predicts the answer $A$ through two stages: (1) structured 3D evidence construction, and (2) reasoning over 3D evidence. Specifically, a question parser first identifies the relevant objects required for answering $Q$, and off-the-shelf perception models are employed to estimate object-level geometric information from $I$, including 3D locations, object orientations, and object sizes. Then, the extracted geometric evidence is converted into a structured textual representation and provided to an LLM, which performs spatial reasoning and generates the final answer.

Formally, the overall process can be formulated as:
\begin{equation}
A = f_{\theta}(E(I), Q),
\end{equation}
where $E(I)$ denotes the structured 3D evidence reconstructed from the image, and $f_{\theta}$ represents the reasoning model parameterized by $\theta$. Unlike end-to-end 
methods
that jointly optimize perception and reasoning, \textit{DiSR} explicitly exposes 
the structured geometric evidence, 
enabling each component to specialize in its corresponding capability.

\subsection{Structured 3D Evidence Construction}
\label{sec:evidence}
To initiate \textit{DiSR}, we reconstruct the physical scene into structured 3D evidence. 
Instead of learning implicit geometric representations within the MLLM, we leverage perception expert models 
specialized 
for 3D geometric estimation.

\noindent{\bf Question Parsing.}
The question parsing identifies the objects required for answering the question and establishes the interface between language understanding and 3D perception. Given a question $Q$, we 
use Qwen3-VL-8B-Instruct as an object planner $f_{\mathrm{plan}}(\cdot)$ to generate an object-centric plan:
\begin{equation}
P=f_{\mathrm{plan}}(Q)
=\{(q_i:m_i)\}_{i=1}^{N},
\label{eq:object_plan}
\end{equation}
where each tuple contains a persistent object slot $q_i$ and its corresponding object reference $m_i$. The slot identifier uniquely tracks the same object throughout the subsequent evidence extraction and reasoning, while the object reference specifies the object to be localized in the image. For example, a question involving a red chair and a table produces:
\begin{equation}
\{
\texttt{q1}: \text{``the red chair''}, \quad \texttt{q2}: \text{``table''}
\}.
\end{equation}

\noindent{\bf Object Grounding.}
Given the object plan $P$, we take Qwen3-VL-8B-Instruct as the grounding module to localize each queried object reference $m_i$ in the image and identify its corresponding region $r_i$. The grounded regions are represented as:
\begin{equation}
R=\{(q_i:r_i)\}_{i=1}^{N},
\end{equation}
where the slot identifier $q_i$ preserves the correspondence between the language query and the visual region. It bridges linguistic references, such as ``the chair on the left'' or ``the object highlighted by the red box'', with localized visual targets. By retaining only the regions specified in $P$, \textit{DiSR} performs target-specific perception and avoids unnecessary computation cost on uninterested objects.

\noindent{\bf Reconstructing the Physical Scene to Derive Evidence.}
Given the grounded regions $R$, \textit{DiSR} uses frozen expert perception models to recover geometric information. Specifically, SAM~\citep{kirillov2023segment} extracts object masks, 
Metric3D~\citep{yin2023metric3d} estimates metric depth, 
and WildCamera~\citep{zhu2023tame} estimates intrinsic parameters and
uses PerspectiveFields~\citep{jin2023perspective} to recover gravity-related camera
orientation.
The depth map is then back-projected and transformed into a shared,
gravity-aligned coordinate frame.
For each grounded mask region, \textit{DiSR} extracts the associated 3D points, removes unreliable outliers, and computes an object-centric 3D
extent and center. 
In addition, OrientAnything~\citep{wang2024orient} estimates object orientations, which are transformed into the same coordinate frame.

The resulting object-centric scene representation is defined as:
\begin{equation}
E(I) =
\{e_i\}_{i=1}^{N},
\quad
e_i=
(\mathbf{c}_i,\mathbf{d}_i,\sigma_i,\rho_i,
\mathbf{f}_i,\mathbf{l}_i),
\label{eq}
\end{equation}
where $\mathbf{c}_i\in\mathbb{R}^{3}$ denotes the object center, $\mathbf{d}_i\in\mathbb{R}^{3}$ denotes its 3D extent, $\sigma_i$ and $\rho_i$ represent object-level size descriptors, 
$\mathbf{f}_i$ and $\mathbf{l}_i$ denote the estimated front and left axes. 
Since all objects are represented in a shared gravity-aligned coordinate system, spatial comparisons can be performed directly from these geometric primitives without requiring the reasoner to interpret raw visual signals, thus eliminating large-scale 3D VQA training to learn implicit 3D geometry.

\subsection{Reasoning over Structured 3D Evidence}
\label{sec:reasoning}
After obtaining structured 3D evidence $E$, \textit{DiSR} employs an LLM with LoRA to perform high-level spatial reasoning. The objective is not to learn geometric perception from raw visual inputs, but to infer spatial relationships and answer questions based on the explicit geometric evidence.

To bridge the perception output and the language model, the structured evidence $E$ is serialized into a textual description with language:
\begin{equation}
E_T = \phi(E(I)),
\end{equation}
where $\phi(\cdot)$ denotes the evidence serialization function. The resulting sequence contains object identities and their geometric attributes, which can be directly interpreted by the LLM.
Given the question $Q$ and serialized evidence $E_T$, the LLM generates the answer:
\begin{equation}
A=f_{\theta}(E_T,Q).
\end{equation}

Since the geometric information is explicitly provided, the LLM only needs to learn reasoning patterns over the structured 3D evidence. We adopt LoRA-based fine-tuning to efficiently adapt the pretrained LLM while preserving its general reasoning capability.

\subsection{Data Collection and Training}
\label{sec:training}

\noindent{\bf Training Data Construction.}
Following previous work~\citep{ma2026spatialreasoner}, we construct the training dataset from images in Open Images~\citep{kuznetsova2018open} by generating spatial reasoning questions and corresponding answers, resulting in a dataset
\(
\mathcal{D}=\{(I,Q,A)\}.
\)
Since \textit{DiSR} performs question parsing for evidence extraction and spatial reasoning as two sequential stages during inference, we decompose each training sample into two complementary supervision tasks. Specifically, the question parsing task learns to predict the parsing result
\(
(Q,P)
\).
The reasoning task learns to generate the answer conditioned on the extracted 3D evidence,
\(
(Q,E_T,A)
\).
This decomposition enables the LLM to jointly learn accurate question parsing and reasoning over explicit geometric evidence. Since images are not required during training, we also construct synthetic data that approximates the distribution of the collected dataset.

\noindent{\bf Parameter-Efficient Fine-tuning.}
We adopt LoRA to efficiently fine-tune the reasoning LLM while keeping all expert perception models frozen. Given the constructed training data, the LLM is optimized using the standard autoregressive language modeling objective:
\begin{equation}
\mathcal{L}
=
-\sum_{t=1}^{|Y|}
\log p_{\theta}(y_t \mid y_{<t}, X),
\end{equation}
where \(X\) denotes the corresponding input prompt (i.e., the parsing or reasoning task), and \(Y\) denotes the target output sequence. During training, only the LoRA parameters are updated, resulting in an efficient adaptation of the pretrained LLM without modifying the perception modules.

\section{Experiments}
\label{sec:exp}

To validate the effectiveness of our \textit{DiSR}, we evaluate our model on: 1) three spatial reasoning benchmarks; 2) five general reasoning benchmarks. In this section, we first introduce our experimental setup~\ref{sec:exp_setup}, then present results on spatial reasoning~\ref{sec:main_results_spatial} and general reasoning benchmarks~\ref{sec:main_results_general}, followed by ablation study~\ref{sec:ablation_study} for disentangled error analysis of perception and reasoning.

\subsection{Experimental Setup}
\label{sec:exp_setup}

\noindent{\bf Evaluation Benchmarks.}
We evaluate \textit{DiSR} on three representative spatial reasoning
benchmarks.
3DSRBench~\citep{ma20253dsrbench} provides a fine-grained evaluation
of 3D spatial reasoning across height, location, orientation, and
multi-object relations.
CV-Bench-3D~\citep{tong2024cambrian} evaluates qualitative depth and
distance reasoning beyond two-dimensional image-plane cues.
SPAR-Bench~\citep{zhang2026flatland} covers tasks of depth estimation, metric
distance estimation, and relational selection.
The numerical tasks are evaluated using mean relative accuracy
(MRA), while the relational-selection tasks are evaluated using
accuracy.

Moreover, we examine whether the fine-tuned \textit{DiSR} models
retain strong general visual-reasoning capabilities on
MMBench~\citep{liu2024mmbench}, GQA~\citep{hudson2019gqa}, POPE~\citep{li2023evaluating},
SEED~\citep{li2023seed}, and RealWorldQA~\citep{xai2024realworldqa}.
Finally, we evaluate object-grounding performance on CV-Bench-3D.

\noindent{\bf Comparison Methods.}
We compare \textit{DiSR} with three groups of representative
baselines:
(1) general-purpose open-source MLLMs, including LLaVA, Cambrian,
and the Qwen-VL family;
(2) proprietary MLLMs, including GPT-4o, Claude, Gemini, and
QwenVLMax; and
(3) spatially specialized models, including SpaceLLaVA, SpatialBot,
SpatialLLM, SpatialRGPT, SpatialReasoner, and HiSpatial.

\noindent{\bf Implementation Details.}
We instantiate \textit{DiSR} with Qwen3-VL-8B-Instruct, denoted as \textit{DiSR-8B-LoRA}. Specifically, for object grounding in deriving 3D evidence, we use both the vision encoder and LLM backbone, while only the LLM is adopted for reasoning and question parsing.

All expert perception modules remain frozen throughout training.
We fine-tune the LLM by applying LoRA with rank $128$, scaling factor
$256$, a learning rate of $2e-5$, and a maximum sequence
length of $1024$. A total of $0.33$M training data is used.
As \textit{DiSR} doesn't need input images during training, a single RTX 4080 SUPER can provide enough memory to support 59 GPU hours of training.

\subsection{Performance on Spatial Reasoning Benchmarks}
\label{sec:main_results_spatial}
\noindent{\bf Results on 3DSRBench.}
Table~\ref{tab:3dsrbench_main} compares 3DSRBench performance.
\textit{DiSR-8B-LoRA} achieves an overall
accuracy of $67.62\%$, significantly outperforming the strongest prior method HiSpatial by $\textbf{3.77\%}$.
We observe that improvements are more pronounced on multi-object reasoning, where \textit{DiSR} models obtain around $63.30\%$, exceeding
HiSpatial by about $\textbf{12.1\%}$ and SpatialReasoner by about
$\textbf{11.5\%}$ individually.
These results indicate that complex reasoning over explicit 3D evidence is
particularly effective for relations that require combining
geometric information across multiple objects.

\begin{table*}[t]
\centering
\caption{
Results on 3DSRBench and efficiency comparisons. Training efficiency is measured by GPU hours. ``NR'' indicates that the number is not reported. 
``*'' represents additional data regarding ``Relational Selection'' task is used.}
\vspace{-0.05in}
\label{tab:3dsrbench_main}

\resizebox{0.93\textwidth}{!}{
\begin{tabular}{lccccccc} 
\toprule
Method
& Overall
& Height
& Location
& Orientation
& Multi-Object
& Spatial Data
& Reported Training \\
\midrule

\multicolumn{8}{l}{\emph{Open-source generalist models}} \\

LLaVA-NeXT-8B
& 48.40 & 50.60 & 59.90 & 36.10 & 43.40
& -- & -- \\

Qwen2.5-VL-7B-Instruct
& 48.40 & 44.10 & 62.70 & 40.60 & 40.50
& -- & -- \\

Qwen3-VL-8B-Instruct
& 53.70 & 48.00 & 71.00 & 40.50 & 46.70
& -- & -- \\

Qwen2.5-VL-72B-Instruct
& 54.90 & 53.30 & 71.00 & 43.10 & 46.60
& -- & -- \\

\midrule
\multicolumn{8}{l}{\emph{Proprietary models}} \\

GPT-4o
& 44.20 & 53.20 & 59.60 & 21.60 & 39.00
& NR &NR \\

Claude-3.5-Sonnet
& 48.20 & 53.50 & 63.10 & 31.40 & 41.30
& NR &NR \\

Gemini-2.0-Flash Thinking
& 51.10 & 53.00 & 67.10 & 35.80 & 43.60
& NR & NR \\

QwenVLMax
& 52.00 & 45.10 & 70.70 & 37.70 & 44.80
& NR & NR \\

\midrule
\multicolumn{8}{l}{\emph{Spatial specialist models}} \\

SpaceLLaVA
& 42.00 & 49.30 & 54.40 & 27.60 & 35.40
& 0.03M & NR \\

SpatialBot
& 41.00 & 40.40 & 54.40 & 31.90 & 33.50
& NR & A100 / 120h \\

SpatialLLM
& 44.80 & 45.80 & 61.60 & 30.00 & 36.70
& 2M & A100 / NR \\

SpatialRGPT w/ depth
& 48.40 & 55.90 & 60.00 & 34.20 & 42.30
& 8.7M & A100 / NR \\

SpatialReasoner
& 60.30 & 52.50 & \textbf{75.20} & 55.20 & 51.80
& 0.05M &H100 / NR \\

HiSpatial-3B (RGB-XYZ)
& 63.85
& \textbf{70.29}
& 66.17
& \textbf{76.76}
& 51.20
& 2B 
& H100 / 1536h \\

\midrule

\textbf{DiSR-8B-LoRA}
& \textbf{67.62}
& \underline{62.75}
& \underline{72.85}
& \underline{69.40}
& \textbf{63.30}
& 0.33M
&\textbf{RTX 4080S / 59h} \\

\textbf{DiSR-8B-LoRA*}
& \underline{65.81}
& 62.61
& 69.81
& 67.86
& \underline{61.90}
& 0.36M
&\textbf{RTX 4080S / 67h} \\

\bottomrule
\end{tabular}
}
\end{table*}

\begin{table*}[t]
\centering
\caption{
Results on SPAR-Bench. ``*'' represents additional data regarding ``Relational Selection'' task is used.
}
\vspace{-0.05in}
\label{tab:spar_single_view}

\resizebox{0.93\textwidth}{!}{
\begin{tabular}{lccccccccc}
\toprule

\multicolumn{1}{c}{\multirow{2}{*}{\textbf{Model}}}
& \multicolumn{5}{c}{\textbf{Metric Estimation (MRA)}}
& \multicolumn{3}{c}{\textbf{Relational Selection (Acc.)}}
& \multirow{2}{*}{\textbf{Avg.}} \\

\cmidrule(lr){2-6}
\cmidrule(lr){7-9}

&
\textbf{Depth-OC}
& \textbf{Depth-OO}
& \textbf{Dist-OC}
& \textbf{Dist-OO}
& \textbf{Metric Avg.}
& \textbf{DistI-OO}
& \textbf{ObjRel-OO}
& \textbf{Rel. Avg.}
& \\

\midrule

Qwen2.5-VL-7B-Instruct
& 32.14
& 20.27
& 42.75
& 25.37
& 30.13
& 58.82
& 52.47
& 55.65
& 38.64 \\

Qwen3-VL-8B-Instruct
& 47.06
& 17.80
& 19.08
& 50.25
& 33.55
& 72.35
& 61.26
& 66.81
& 44.63 \\

SpatialReasoner
& 29.86
& 17.77
& 21.40
& 40.55
& 27.40
& 15.29
& 16.21
& 15.75
& 23.51 \\

HiSpatial-3B (RGB-XYZ)
& 41.41
& 14.15
& 7.68
& \textbf{53.17}
& 29.10
& \textbf{81.18}
& 59.07
& \textbf{70.12}
& 42.78 \\

\midrule

\textbf{DiSR-8B-LoRA}
& 57.20
& 27.35
& \textbf{67.66}
& 45.26
& 49.37
& 54.12
& 26.37
& 40.25
& 46.33 \\

\textbf{DiSR-8B-LoRA*}
& \textbf{64.82} 
& \textbf{27.37} 
& 63.29
& \textbf{53.27} 
& \textbf{52.19}
& 67.65 
& \textbf{68.96} 
& 68.30 
& \textbf{57.56} \\
\bottomrule
\end{tabular}
}
\end{table*}

\noindent{\bf Results on SPAR-Bench.}
Table~\ref{tab:spar_single_view} reports performance comparisons on SPAR-Bench. \textit{DiSR-8B-LoRA} achieves the best overall score of
$46.33\%$, outperforming Qwen3-VL-8B-Instruct by $\textbf{1.70\%}$ and the strongest prior spatial model HiSpatial by $\textbf{3.55\%}$.
An interesting phenomenon is observed: \textit{our \textit{DiSR} achieves 49.37\% for metric estimation tasks, significantly surpassing Qwen3-VL-8B-Instruct and HiSpatial by \textbf{15.82\%} and \textbf{20.27\%}, respectively. However, \textit{DiSR} performs poorly on relational selection tasks.}
The reason behind this is that we follow SpatialReasoner for the training data construction, and the generated 0.33M training data can not cover the data distribution required for the relational selection tasks.
With additional 2000 data regarding ``Relatinal Selection'' task, the performance of \textit{DiSR} is significantly enhanced, surpassing the previous best model by \textbf{12.93\%} overall score.

\noindent{\bf Results on CV-Bench-3D.}
Table~\ref{tab:cvbench3d} reports the results on CV-Bench-3D. \textit{DiSR} only achieves comparable performance to Qwen3-VL-8B-Instruct. With a deep analysis, we observe the two important reasons: (I) as identified by previous work~\cite{ma2026spatialreasoner}, CV-Bench-3D contains substantial 2D shortcuts, which allow end-to-end MLLMs to infer answers without knowing the 3D geometry. (II) CV-Bench-3D suffers from low-quality images. It is challenging for current perception models to reconstruct accurate 3D evidence for small objects in low-quality images. 
Please refer to the Appendix for more detailed analysis.

The contrast between CV-Bench-3D accuracy in Table~\ref{tab:cvbench3d} and the grounding results in Table~\ref{tab:grounding} shows a surprising phenomenon: \textit{while HiSpatial achieves strong performance on CV-Bench-3D, it cannot even infer reliable object localization, leaving a mystery about how spatial reasoning performs in the model}. In contrast, \textit{DiSR} explicitly reconstructs structured 3D evidence and performs reasoning over this intermediate representation, making the geometric reasoning process inspectable and allowing errors to be attributed to grounding, 3D modeling, or reasoning.

In interactive applications, users can directly specify the queried objects instead of relying on model-predicted grounding. To simulate this setting, we replace the predicted object boxes with ground-truth boxes while keeping the same DiSR-8B-LoRA checkpoint and reasoning pipeline. As shown in Table~\ref{tab:cvbench3d}, the average accuracy increases from 92.25\% to 93.75\%, with improvements of 1.34\% and 1.66\% on the Depth and Distance tasks, respectively. This controlled improvement identifies object grounding as a remaining performance bottleneck and demonstrates that stronger grounding models or user-provided object specifications can be incorporated without modifying the reasoning module.

\noindent{\bf Training Efficiency Discussion.}
\textit{DiSR} is highly training-efficient. Specifically, it only fine-tunes the LoRA parameters using 0.33M training samples on a single RTX 4080 SUPER GPU for 59 hours. In contrast, HiSpatial fine-tunes the entire MLLM on 2 billion spatial QA pairs using 32$\times$48 H100 GPU hours. Despite this substantial reduction in training data, trainable parameters, and computational cost, \textit{DiSR} achieves superior spatial reasoning performance. These results suggest that strong spatial reasoning does not necessarily require jointly learning 3D perception and reasoning through large-scale spatial supervision. Instead, by explicitly disentangling the two, specialized perception models can reconstruct reliable geometric evidence, allowing the LLM to focus on high-level compositional reasoning over structured 3D representations.

\begin{table}[t]
\centering
\caption{
Results on CV-Bench-3D.
``w/ GT grounding'' replaces model-predicted object boxes with
ground-truth boxes to simulate user-specified targets.
}
\vspace{-0.05in}
\label{tab:cvbench3d}

\resizebox{0.47\textwidth}{!}{%
\begin{tabular}{lccc}
\toprule
Method
& Depth $\uparrow$
& Distance $\uparrow$
& Avg. $\uparrow$ \\
\midrule
Qwen2.5-VL-7B-Instruct
& 84.50
& 80.00
& 82.25 \\

Qwen3-VL-8B-Instruct
& 95.50
& 89.83
& 92.67 \\

SpatialReasoner
& 87.00
& 72.83
& 79.92 \\

HiSpatial-3B (RGB)
& -
& -
& 95.58 \\

\midrule
DiSR-8B-LoRA
& 92.83
& 91.67
& 92.25 \\

DiSR-8B-LoRA w/ GT grounding
& 94.17
& 93.33
& 93.75 \\
\bottomrule
\end{tabular}
}
\end{table}

\subsection{Performance on General Reasoning Benchmarks}
\label{sec:main_results_general}
Spatial reasoning fine-tuning might improve domain-specific performance at the expense of the general visual understanding capabilities inherited from the pretrained MLLM. To evaluate this trade-off, we assess \textit{DiSR} on five general visual reasoning benchmarks. As shown in Table~\ref{tab:general_vqa} , \textit{DiSR} consistently achieves performance comparable to its base model, \emph{i.e.}, Qwen3-VL-8B-Instruct, indicating that the substantial gains on spatial reasoning are achieved while preserving the general reasoning capabilities of the pretrained MLLMs.

In contrast, prior methods exhibit less favorable trade-offs. SpatialReasoner suffers a noticeable drop in general visual reasoning after fine-tuning on a relatively small 3D VQA dataset. HiSpatial is optimized on large-scale 3D VQA data and even improves over its base MLLM on general visual reasoning benchmarks. However, as shown in Table~\ref{tab:grounding}, both HiSpatial and SpatialReasoner experience a substantial decline in object grounding accuracy. These results demonstrate that the explicit disentanglement of perception and reasoning enables \textit{DiSR} to specialize in spatial reasoning while retaining the broad visual reasoning capabilities inherited from the pretrained MLLM.

\begin{table}[t]
\centering
\caption{
Results on general multimodal reasoning benchmarks.
}
\vspace{-0.05in}
\label{tab:general_vqa}

\resizebox{0.47\textwidth}{!}{
\begin{tabular}{lccccc} 
\toprule
Model
& MMBench
& GQA
& POPE
& SEED
& RealWorldQA \\
\midrule

HiSpatial-3B (RGB-XYZ)
& 69.67
& 61.50
& 87.97
& 63.51
& 58.95 \\

SpatialReasoner
& 79.38
& 61.80
& 85.54
& 73.87
& 66.14 \\
\midrule

Qwen2.5-VL-7B-Instruct
& 84.02
& 61.23
& 87.61
& 77.64
& 68.76 \\

Qwen3-VL-8B-Instruct
& 84.71
& 61.44
& 88.77
& 78.55
& 69.02 \\

\midrule

DiSR-8B-LoRA
& 84.62
& 61.20
& 88.72
& 78.95
& 68.76 \\
\bottomrule
\end{tabular}
}
\end{table}

\subsection{Ablation Study}
\label{sec:ablation_study}

\begin{table}[t]
\centering
\caption{
Interpretability and diagnosability of DiSR on CV-Bench-3D.
}
\vspace{-0.05in}
\label{tab:cvbench3d_analysis}

\resizebox{0.47\textwidth}{!}{%
\begin{tabular}{lcc}
\toprule
Method & Depth & Distance \\

\midrule
DiSR-8B-LoRA & 92.83 & 91.67 \\
DiSR-8B-LoRA w/ gt of question parsing & 92.33 & 92.83 \\
DiSR-8B-LoRA w/ gt of object grounding & 94.17 & 93.33 \\
DiSR-8B-LoRA w/ gt of required 3D evidence& \textbf{100.0}  & \textbf{100.0}  \\

\bottomrule
\end{tabular}
}
\end{table}

\noindent{\bf Interpretability and Diagnosability of \textit{DiSR}.}
By explicitly disentangling perception from reasoning, \textit{DiSR} provides an interpretable reasoning pipeline in which errors can be attributed to individual components rather than an opaque end-to-end model. Since CV-Bench-3D provides ground-truth annotations for the intermediate 3D information required by \textit{DiSR}, we conduct a diagnosability analysis to identify the performance bottlenecks.

Table~\ref{tab:cvbench3d_analysis} presents the results. Replacing the predicted question parsing with ground truth yields only marginal improvements, indicating that question parsing is not the primary source of error. In contrast, using ground-truth object grounding improves the accuracy of the Depth and Distance tasks by 1.34\% and 1.66\%, respectively, suggesting that more accurate object grounding—potentially enabled by stronger MLLMs—would further improve overall spatial reasoning performance. Finally, when the required 3D information is replaced with ground truth, \textit{DiSR} achieves nearly 100\% accuracy on both Depth and Distance, demonstrating that the reasoning module can reliably infer spatial relationships once provided with accurate 3D evidence.

\noindent{\bf Object Grounding Analysis.}
Table~\ref{tab:grounding} reports category-conditioned object grounding on 8,548 unambiguous examples constructed from COCO val2017~\cite{lin2014microsoft}, retaining only image--category pairs with exactly one valid instance. Although \textit{DiSR} fine-tunes the LLM with LoRA for question parsing and reasoning, DiSR-8B-LoRA preserves and slightly improves the grounding capability of Qwen3-VL-8B, achieving higher valid-box coverage and localization accuracy under the same semantic grounding instruction. In contrast, SpatialReasoner exhibits substantially lower IoU despite returning valid boxes for most examples while HiSpatial even fails to return valid boxes. This discrepancy suggests that strong downstream spatial reasoning does not necessarily imply reliable object grounding, while the lightweight adaptation used by \textit{DiSR} retains the visual localization interface required for structured evidence construction.

\begin{table}[t]
\centering
\caption{
Object grounding on the 8,548-example
COCO val2017 unique-instance subset.
\emph{Found} is the percentage of examples for which a model
returns a valid bounding box.
All predictions are converted to original-image pixel coordinates
before evaluation.
}
\vspace{-0.05in}
\label{tab:grounding}

\resizebox{0.47\textwidth}{!}{%
\begin{tabular}{lcccc}
\toprule
Model
& Found $\uparrow$
& mIoU $\uparrow$
& IoU@0.50 $\uparrow$
& IoU@0.75 $\uparrow$ \\
\midrule
Qwen2.5-VL-7B-Instruct
& 96.64
& 64.02
& 71.64
& 54.59 \\

Qwen3-VL-8B-Instruct
& 89.67
& 71.05
& 78.81
& 66.39 \\

\midrule
SpatialReasoner
& 93.66
& 34.82
& 40.51
& 7.07 \\

HiSpatial-3B (RGB-XYZ)
& fails
& --
& --
& -- \\

\midrule

DiSR-8B-LoRA
& \textbf{99.15}
& \textbf{73.99}
& \textbf{81.84}
& \textbf{68.17} \\
\bottomrule
\end{tabular}
}
\end{table}

\noindent{\bf Cross-backbone Evaluation.}
We evaluate the extensibility of \textit{DiSR} by instantiating it with different backbone MLLMs. Besides Qwen3-VL-8B-Instruct, we adopt Qwen2.5-VL-7B-Instruct as the base MLLM. As shown in Table~\ref{tab:extensibility_disr}, \textit{DiSR} consistently achieves strong spatial reasoning performance across both backbones, demonstrating that the framework can seamlessly incorporate advances in foundation models without modifying its overall design. Moreover, a stronger backbone further improves spatial reasoning performance, primarily due to its enhanced general reasoning ability (Table~\ref{tab:general_vqa}) and more accurate object grounding capability (Table~\ref{tab:grounding}). These results highlight the modularity of \textit{DiSR}, where improvements in the underlying MLLM can be directly translated into stronger spatial reasoning performance.

\begin{table}[h]
    \centering
    \caption{Extensibility of \textit{DiSR}.}
    \vspace{-0.05in}
    \resizebox{0.47\textwidth}{!}{%
    \begin{tabular}{cccccc}
    \toprule
     Base MLLM &Height &Location &Orientation &Multi-Object &Overall  \\
     \midrule 
     Qwen2.5-VL-7B-Instruct  &60.43 &71.80 &64.96 &61.67 &65.52 \\
     Qwen3-VL-8B-Instruct    &62.75 &72.85 &69.40 &63.30 &67.62\\
     \bottomrule
    \end{tabular}
     }
    \label{tab:extensibility_disr}
\end{table}

\section{Conclusion}
\label{sec:con}

We presented \textit{DiSR}, a simple yet effective framework that explicitly disentangles 3D perception from spatial reasoning through structured 3D evidence. By leveraging specialized perception models to reconstruct the physical world into reliable 3D evidence and then training an LLM with LoRA to reason solely over explicit geometric evidence, \textit{DiSR} achieves state-of-the-art performance on multiple spatial reasoning benchmarks while requiring substantially less training data, fewer trainable parameters, and significantly lower computational cost than existing approaches. Beyond its strong empirical performance, the explicit separation of perception and reasoning improves interpretability, diagnosability, and modularity, demonstrating that disentangling perception and reasoning is a scalable and effective alternative paradigm for spatial intelligence. 

\raggedbottom
\bibliography{aaai2027}


\setcounter{secnumdepth}{1}
\renewcommand{\thesection}{\Alph{section}}
\setcounter{table}{0}  
\renewcommand{\thetable}{A\arabic{table}}
\renewcommand{\thefigure}{A\arabic{figure}}

\appendix
\section{Additional Results on 3DSRBench}
\label{app:3dsrbench}
\begin{table*}[htb!]
\centering
\caption{
Fine-grained relation accuracy (\%) on 3DSRBench.
The best and second-best results in each row are shown in
\textbf{bold} and \underline{underlined}, respectively.
}
\label{tab:3dsrbench_fine}

\resizebox{0.9\textwidth}{!}{
\begin{tabular}{llcccc} 
\toprule
Category
& Relation
& SpatialReasoner
& HiSpatial-3B
& DiSR-Qwen2.5
& DiSR-Qwen3 \\
\midrule

Height
& Higher
& 51.70
& \textbf{70.29}
& 60.43
& \underline{62.75} \\

\midrule
Location
& Above
& \textbf{70.70}
& 47.71
& 69.36
& \underline{70.38} \\

& Closer to camera
& 80.10
& \textbf{86.29}
& 76.70
& \underline{80.68} \\

& Next to
& \textbf{75.50}
& 62.86
& \underline{66.96}
& 62.24 \\

\midrule
Orientation
& In front of
& 63.70
& \textbf{77.71}
& 71.22
& \underline{75.29} \\

& On the left
& 64.20
& \textbf{85.14}
& 70.20
& \underline{77.08} \\

& Viewpoint
& 36.40
& \textbf{67.43}
& 53.35
& \underline{55.69} \\

\midrule
Multi-Object
& Closer to
& 54.90
& 73.14
& \underline{75.14}
& \textbf{77.14} \\

& Facing
& \underline{68.20}
& 52.57
& 67.34
& \textbf{70.81} \\

& Viewpoint toward object
& 34.40
& 32.57
& \underline{46.94}
& \textbf{50.73} \\

& Parallel
& 50.10
& 47.43
& \textbf{62.54}
& \underline{57.82} \\

& Same direction
& 55.50
& 50.29
& \underline{56.10}
& \textbf{59.59} \\

\bottomrule
\end{tabular}
}
\end{table*}

\paragraph{Fine-grained Relation Analysis.}
Table~\ref{tab:3dsrbench_fine} presents fine-grained comparisons across the 12 relation types in 3DSRBench. HiSpatial achieves strong performance on relations involving height, camera depth, and single-object orientation, including \emph{Higher}, \emph{Closer to camera}, \emph{In front of}, \emph{On the left}, and \emph{Viewpoint}, benefiting from its large-scale spatial training. In contrast, \textit{DiSR} consistently excels at multi-object spatial relations, achieving the best performance on all five such categories. Specifically, DiSR-Qwen3 leads on \emph{Closer to}, \emph{Facing}, \emph{Viewpoint toward object}, and \emph{Same direction}, while DiSR-Qwen2.5 performs best on \emph{Parallel}. These relations require aggregating geometric information across multiple objects or reasoning in object-dependent coordinate systems, which is challenging to learn solely from images. The superior performance of \textit{DiSR} on these tasks highlights the advantage of explicit 3D evidence for reliable and compositional spatial reasoning.

\section{Shortcut and Error Analysis on CV-Bench-3D}
\label{app:cvbench}

Although CV-Bench-3D is designed to evaluate spatial reasoning beyond two-dimensional image-plane cues, its Distance and Depth subsets still contain strong correlations between simple 2D box geometry and the ground-truth answers. To better understand these effects, we first analyze task-specific 2D heuristics for each subset and then exploit the explicit intermediate representations in \textit{DiSR} to identify the remaining sources of error.

\subsection{Distance: Image-Plane Proximity}
\label{app:cvbench_distance}

The Distance subset evaluates whether a model can determine which of two candidate objects is closer to a reference object in 3D space. However, in many samples, the object that is closer to the reference object in 3D also exhibits a smaller 2D center distance. As a result, models may achieve high accuracy by exploiting 2D spatial correlations rather than recovering the underlying 3D relationship, making the reported performance on this subset an imperfect measure of true 3D reasoning ability.

\paragraph{2D Proximity Heuristic.}
Following the analysis of SpatialReasoner,
we consider a simple heuristic that uses neither depth nor reconstructed
3D geometry. Let $\mathbf{c}_{r}$ denote the 2D center of the red
reference box, and let $\mathbf{c}_{b}$ and $\mathbf{c}_{g}$ denote the
centers of the blue and green candidate boxes, respectively. The
heuristic selects the candidate with the smaller image-plane distance
to the reference:
\begin{equation}
\hat{y}_{\mathrm{2D}}
=
\operatorname*{arg\,min}_{o\in\{b,g\}}
\left\lVert
\mathbf{c}_{r}-\mathbf{c}_{o}
\right\rVert_{2}.
\label{eq:distance_2d_shortcut}
\end{equation}
Despite its simplicity, this rule obtains \textbf{80.17\%} accuracy on the 600 Distance questions of CV-Bench-3D. In contrast, it obtains only \textbf{34.30\%} on the \emph{``Multi-Object Closer to"} relation of
3DSRBench, where image-plane proximity is substantially less predictive of the annotated 3D relationship.

\paragraph{2D-Consistent and 2D-Conflict Subsets.}
To separate samples where image-plane cues correlate with the underlying 3D relation from those where they fail, we partition the 600 Distance questions into two disjoint subsets. The \emph{2D-consistent} subset consists of 481 samples where the 2D heuristic prediction, $\hat{y}_{\mathrm{2D}}$, agrees with the ground-truth answer, whereas the remaining 119 samples form the \emph{2D-conflict} subset where the heuristic produces the opposite prediction. We also report the accuracy for the two subsets.

\begin{table*}[t]
\centering

\caption{
Analysis of distance-related reasoning on CV-Bench-3D and
3DSRBench. The CV-Bench-3D Distance subset is divided according
to whether the 2D bounding-box-center heuristic agrees with the
ground-truth answer. The final column reports accuracy on the
corresponding \emph{Multi-Object Closer to} relation of 3DSRBench.
}

\small
\setlength{\tabcolsep}{7.0pt}
\begin{tabular}{lcccc}
\toprule
\multicolumn{1}{c}{\multirow{2}{*}{Method}}
& \multicolumn{3}{c}{CV-Bench-3D Distance}
& 3DSRBench \\
\cmidrule(lr){2-4}
& Overall
& 2D-Consistent (481)
& 2D-Conflict (119)
& Multi-Object Closer to \\
\midrule
2D Heuristic
& 80.17
& 100.00
& 0.00
& 34.30 \\

SpatialReasoner
& 72.83
& 75.26
& 63.03
& 54.90 \\

\midrule
Qwen3-VL-8B-Instruct
& 89.83
& 98.96
& 52.94
& 61.40 \\

DiSR-8B-LoRA
& \textbf{91.67}
& 95.84
& \textbf{74.79}
& \textbf{77.14} \\
\bottomrule
\end{tabular}

\label{tab:cvbench3d_distance_combined}
\end{table*}

\noindent{\bf Discussion on Results.}
Table~\ref{tab:cvbench3d_distance_combined} jointly reveals a
cross-benchmark discrepancy and substantial shortcut dependence
within CV-Bench-3D Distance. The 2D heuristic achieves 80.17\%
accuracy on CV-Bench-3D but only 34.30\% on the corresponding
3DSRBench, indicating that image-plane proximity is
substantially more predictive on CV-Bench-3D.
Qwen3-VL-8B-Instruct similarly drops from 98.96\% on the
2D-consistent subset to 52.94\% on the 2D-conflict subset. 
In contrast, DiSR-8B-LoRA improves conflict
accuracy to 74.79\% and consistently achieves the strongest
result, 77.14\%, on the corresponding 3DSRBench. These
results indicate that reasoning over explicit 3D evidence is a more reliable alternative for spatial intelligence.

\begin{figure*}[t]
\centering
\includegraphics[width=1.0\textwidth]
{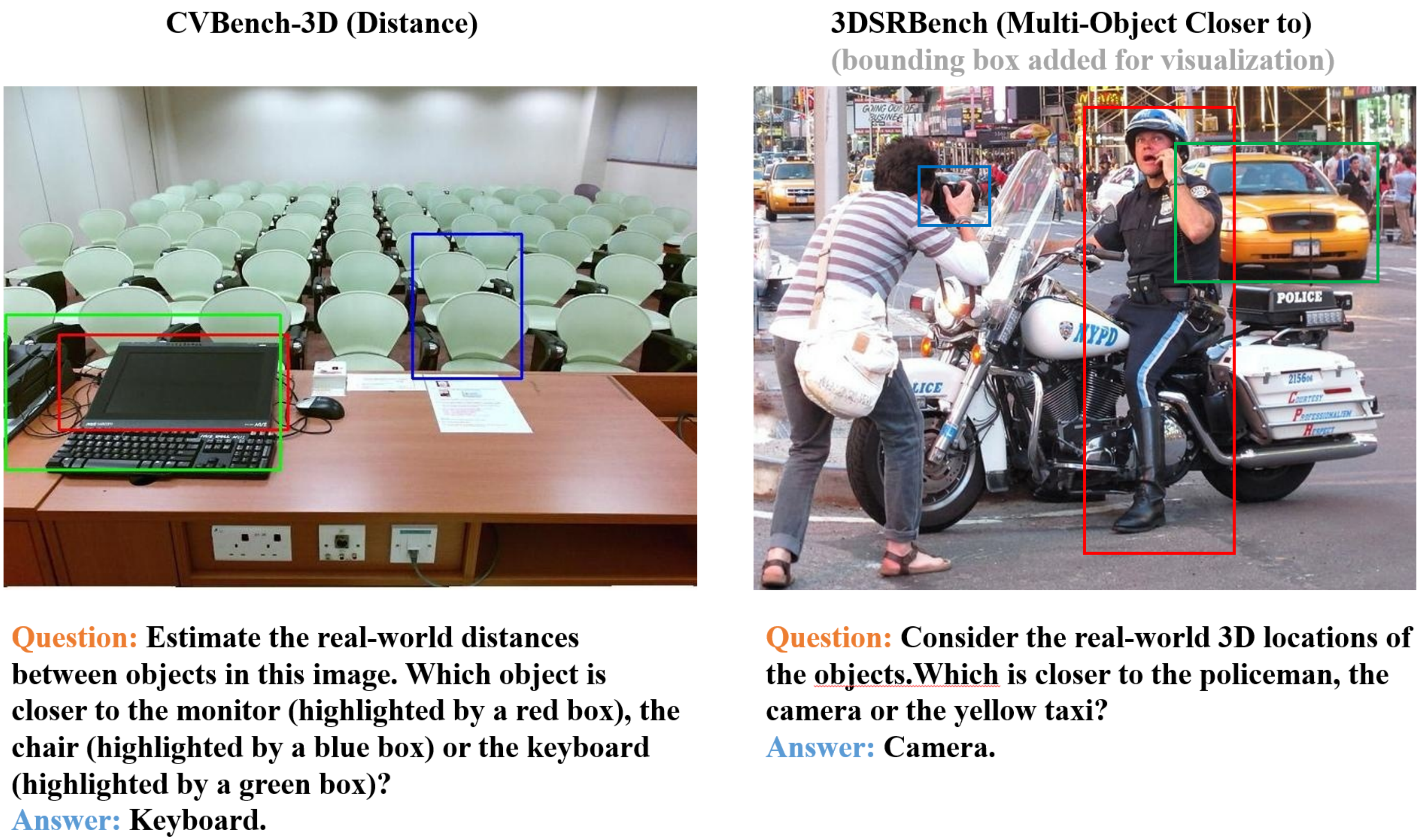}
\caption{
Qualitative examples illustrating the 2D shortcut in CV-Bench-3D Distance and its failure to transfer reliably to 3DSRBench.
\textbf{Left:} In the CV-Bench-3D Distance example, the keyboard
occupies an image-plane region close to and partially overlapping the
red reference box around the monitor, whereas the chair is projected
substantially farther away. The 2D proximity heuristic therefore
selects the keyboard, matching the annotated answer as well as the
predictions of Qwen3-VL-8B-Instruct and DiSR-8B-LoRA.
\textbf{Right:} In the 3DSRBench \emph{Multi-Object Closer to}
example, the center of the yellow-taxi box lies closer to the
police-officer box in the image plane than the center of the camera
box. Nevertheless, the annotated 3D answer is the camera,
demonstrating that image-plane proximity is not a reliable proxy for
the underlying 3D relation. Bounding boxes in the 3DSRBench image are
added only for visualization and are not part of the original
benchmark input.
}
\label{fig:distance_shortcut_examples}
\end{figure*}

Figure~\ref{fig:distance_shortcut_examples} provides qualitative examples that complement the quantitative analysis. In the CV-Bench-3D example, the correct answer can be inferred almost directly from the relative positions of the highlighted 2D bounding boxes, indicating the presence of strong image-plane cues. In contrast, applying the same 2D heuristic to the 3DSRBench example leads to an incorrect prediction, despite all queried objects being clearly visible. This discrepancy highlights the difference in shortcut availability across benchmarks and motivates our further analysis of 2D shortcut dependence in CV-Bench-3D.

\subsection{Depth: Bottom-Position Shortcut}
\label{app:cvbench_depth}

The Depth subset evaluates whether a model can determine which of two
highlighted objects is closer to the camera. Although this task
requires camera-depth reasoning, the annotated depth relation is
strongly correlated with the vertical positions of the object boxes.
In many ground-plane scenes, the object closer to the camera is
projected lower in the image, creating a potentially strong
image-plane shortcut.

\paragraph{2D Bottom-Position Heuristic.}
Let $y_{r}^{\mathrm{bot}}$ and $y_{b}^{\mathrm{bot}}$ denote the
bottom coordinates of the red and blue object boxes, respectively,
where the image $y$-coordinate increases downward. We define a simple
2D heuristic that selects the object with the larger bounding-box-bottom
coordinate, corresponding to a lower position in the image:

\begin{equation}
\hat{y}_{\mathrm{bottom}}
=
\operatorname*{arg\,max}_{o\in\{r,b\}}
y_{o}^{\mathrm{bot}}.
\label{eq:depth_bottom_shortcut}
\end{equation}

Despite using neither depth information nor reconstructed 3D geometry,
this heuristic correctly answers 512 of the 600 Depth questions,
achieving \textbf{85.33\%} accuracy. We therefore partition the
samples into 512 \emph{2D-consistent} cases, for which the heuristic
agrees with the ground-truth answer, and 88 \emph{2D-conflict} cases,
for which it selects the incorrect object.

\begin{figure*}[!t]
\centering
\includegraphics[width=1.0\textwidth]
{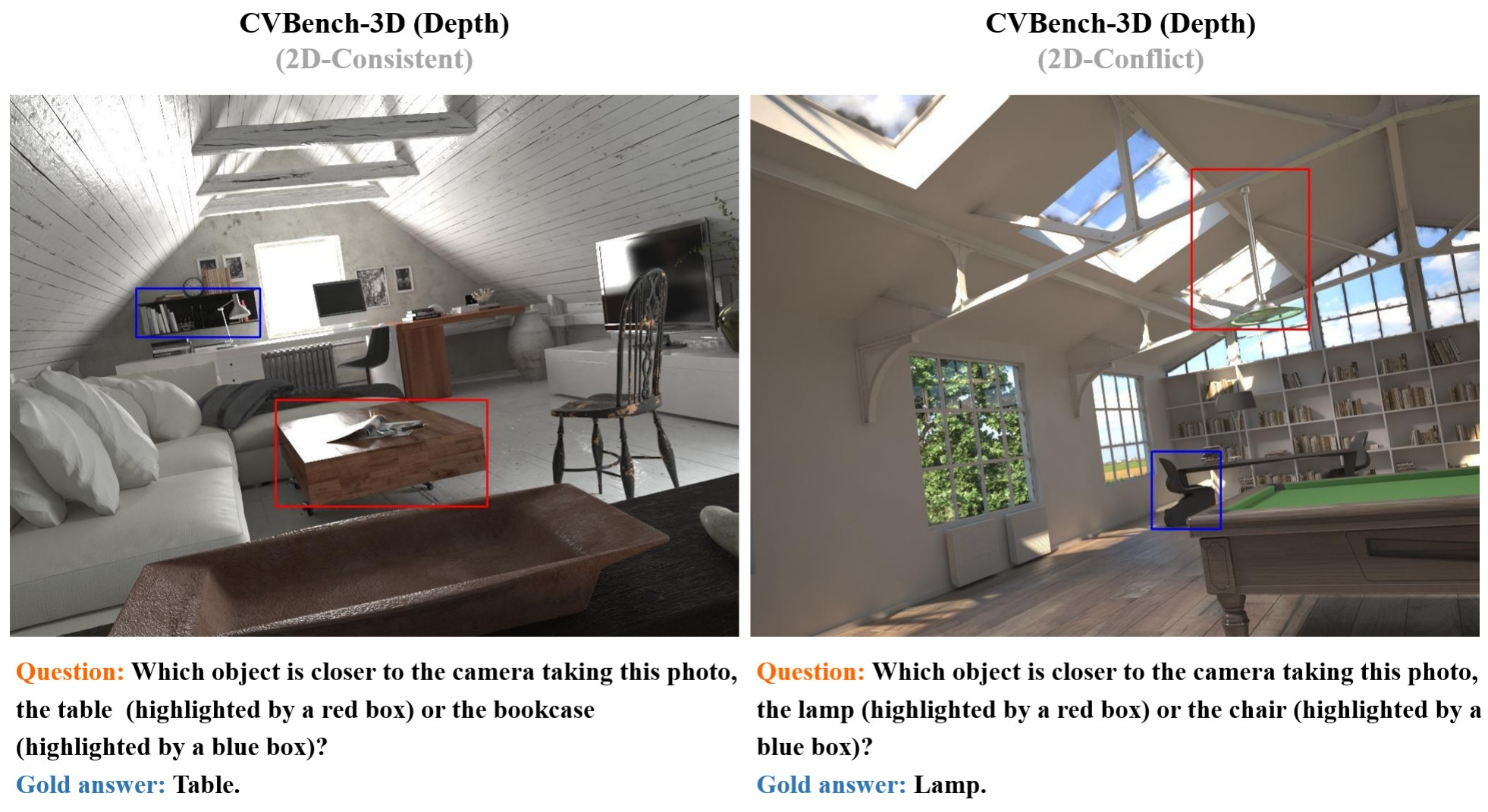}
\caption{
Representative examples from the 2D-consistent and
2D-conflict subsets of CV-Bench-3D Depth analyzed in
Table~\ref{tab:cvbench3d_depth_bottom_split}.
\textbf{Left:} In the 2D-consistent example, the table has a lower
bounding-box-bottom position than the bookcase and is also annotated
as the object closer to the camera. The bottom-position heuristic
therefore produces the correct answer without using explicit depth
information.
\textbf{Right:} In the 2D-conflict example, the chair has a lower
bounding-box-bottom position than the hanging lamp, causing the same
heuristic to select the chair, whereas the annotated closer object is
the lamp. These examples illustrate when the bottom-position shortcut
agrees with the underlying camera-depth relation and when it becomes
misleading.
}
\label{fig:depth_bottom_shortcut_example}
\end{figure*}

Figure~\ref{fig:depth_bottom_shortcut_example} provides representative
examples from the two subsets. In the \emph{2D-consistent} example,
the table has a lower bounding-box-bottom position than the bookcase
and is also annotated as the object closer to the camera. The
bottom-position heuristic therefore produces the correct answer
without using explicit depth information. In the \emph{2D-conflict}
example, the chair has a lower bounding-box-bottom position than the
hanging lamp, causing the same heuristic to select the chair, whereas
the annotated closer object is the lamp. These examples illustrate
both the strong predictive correlation captured by the heuristic and
the cases in which this image-plane cue conflicts with the underlying
camera-depth relation.

\begin{table*}[t]
\centering

\caption{
Accuracy (\%) on the CV-Bench-3D Depth subset after dividing the
samples according to whether the 2D bounding-box bottom-position
heuristic agrees with the ground-truth answer. The heuristic obtains
100.00\% on the 2D-consistent subset and 0.00\% on the 2D-conflict
subset by construction. The best result among the learned models in
each column is shown in \textbf{bold}.
}

\small
\setlength{\tabcolsep}{13pt}
\begin{tabular}{lccc}
\toprule
\multicolumn{1}{c}{Method}
& Overall
& 2D-Consistent (512)
& 2D-Conflict (88) \\
\midrule
2D Heuristic
& 85.33
& 100.00
& 0.00 \\

SpatialReasoner
& 87.00
& 87.30
& 85.23 \\
\midrule
Qwen3-VL-8B-Instruct
& \textbf{95.50}
& \textbf{96.09}
& \textbf{92.05} \\

DiSR-8B-LoRA
& 92.83
& 93.55
& 88.64 \\
\bottomrule
\end{tabular}

\label{tab:cvbench3d_depth_bottom_split}
\end{table*}

\noindent{\bf Discussion on Results.}
Table~\ref{tab:cvbench3d_depth_bottom_split} confirms that
CV-Bench-3D Depth contains a strong bottom-position shortcut. The
simple 2D heuristic correctly answers 512 of the 600 questions,
indicating that the annotated closer object is usually projected lower
in the image. Figure~\ref{fig:depth_bottom_shortcut_example}
qualitatively illustrates both sides of this correlation: the cue
directly yields the correct answer in a 2D-consistent sample but becomes
misleading in a 2D-conflict sample.

Qwen3-VL-8B-Instruct and DiSR-8B-LoRA retain 92.05\% and 88.64\%
accuracy, respectively, on the 2D-conflict subset, showing that both
models can often recover the correct camera-depth relation even when
the bottom-position cue is misleading. Moreover,
Qwen3-VL-8B-Instruct outperforms DiSR-8B-LoRA on both the
2D-consistent and 2D-conflict subsets. Its overall advantage therefore
cannot be explained solely by stronger exploitation of the
bottom-position shortcut. The object-size analysis in
Table~\ref{tab:depth_object_size_analysis} instead shows that the
remaining errors of DiSR-8B-LoRA are concentrated in samples
containing small target objects, for which automatically constructed
object-level 3D evidence is less reliable.

\begin{table*}[t]
\centering

\caption{
Depth accuracy (\%) stratified by the image-plane size of the
ground-truth answer object. The 600 samples are divided into five
equally sized groups of 120 samples. ``DiSR w/ GT Object Grounding''
uses ground-truth object regions while retaining automatically
constructed 3D evidence. ``DiSR w/ GT 3D Evidence'' additionally
replaces the automatically constructed evidence with ground-truth
3D evidence, achieving 100\% accuracy in every size group.
}

\small
\setlength{\tabcolsep}{7pt}
\begin{tabular}{lccccc}
\toprule
Answer-object size
& Median box
& Qwen3-VL-8B
& DiSR-8B
& DiSR w/ GT Object
& DiSR w/ GT 3D \\
& area (\%)
& Instruct
& LoRA
& Grounding
& Evidence \\
\midrule
Smallest 20\%
& 0.236 & 89.17 & 84.17 & 85.00 & 100.00 \\

Small
& 0.786 & 97.50 & 95.00 & 95.00 & 100.00 \\

Medium
& 2.184 & 95.83 & 92.50 & 93.33 & 100.00 \\

Large
& 5.618 & 96.67 & 94.17 & 95.00 & 100.00 \\

Largest 20\%
& 14.453 & 98.33 & 98.33 & 98.33 & 100.00 \\
\bottomrule
\end{tabular}

\label{tab:depth_object_size_analysis}
\end{table*}

\subsection{Structured 3D Evidence for Small Objects}

Table~5 in the main paper identifies that the performance bottleneck of \textit{DiSR} on CV-Bench-3D lies in the inaccurate 3D evidence.
As we observe that the images in CV-Bench-3D are often in low-resolution,
we further investigate whether structured 3D evidence will
become less reliable when the queried object occupies only a small
image region. For each question, we measure the image-plane size of
the ground-truth answer object using its annotated 2D bounding box:

\begin{equation}
s_{\mathrm{ans}}
=
\frac{w_{\mathrm{ans}}h_{\mathrm{ans}}}{WH},
\label{eq:answer_object_scale}
\end{equation}

where $W$ and $H$ denote the image width and height, respectively. The 600 Depth
samples are sorted according to $s_{\mathrm{ans}}$ and divided into
five equally sized groups of 120 samples, ranging from the smallest to
the largest answer objects.

\begin{figure*}[!t]
\centering
\includegraphics[width=0.96\textwidth]
{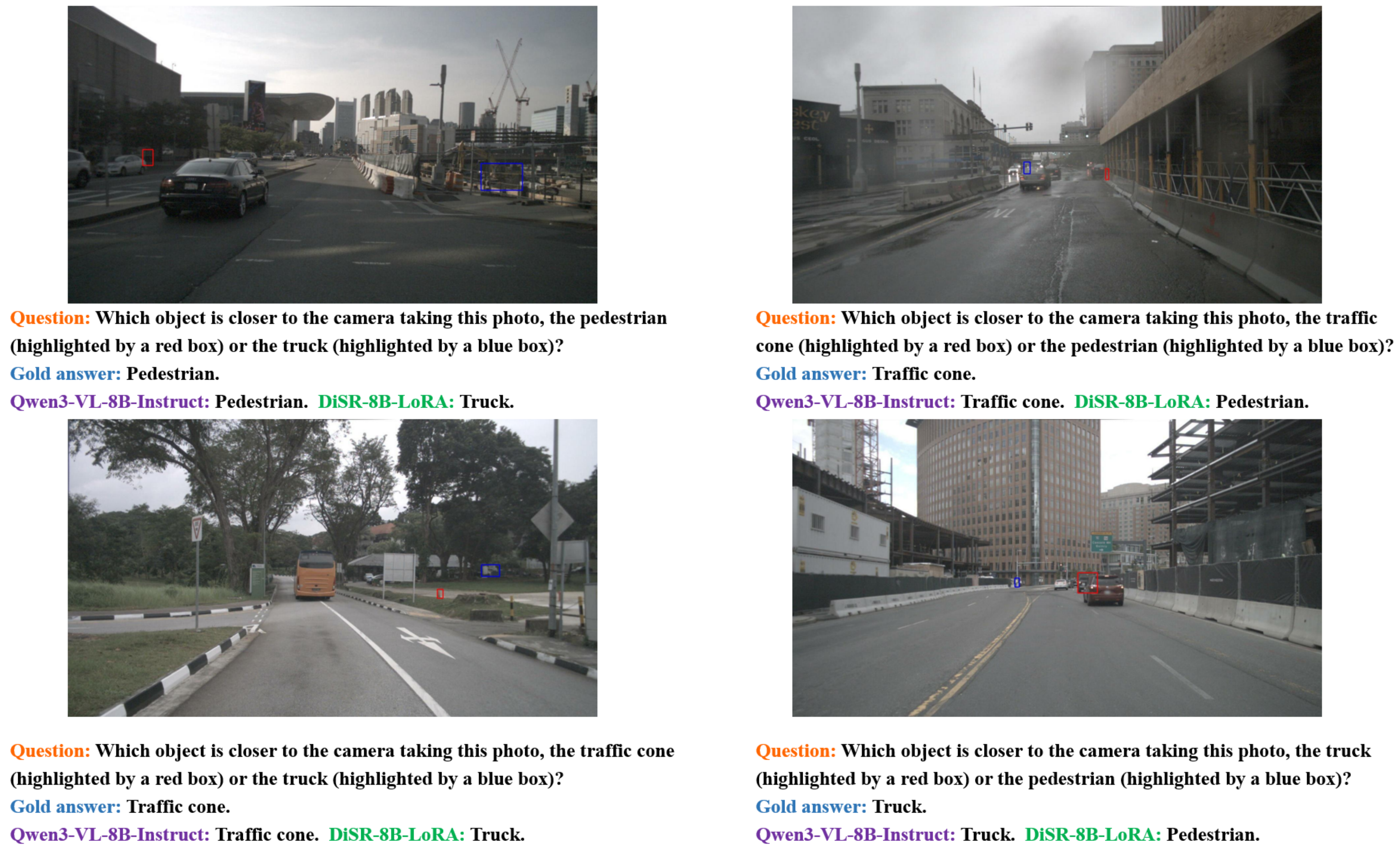}
\caption{
Four representative failure cases involving small image-plane target objects from CV-Bench-3D Depth, corresponding to the size-stratified analysis in Table~\ref{tab:depth_object_size_analysis}. Across these examples, Qwen3-VL-8B-Instruct produces the correct answer, whereas DiSR-8B-LoRA makes an incorrect prediction. All four examples are selected from samples that remain incorrect under GT Object Grounding but are recovered with GT 3D Evidence, indicating errors in the automatically constructed object-level geometry.
}
\label{fig:depth_small_object_evidence_example}
\end{figure*}

\paragraph{Discussion on Results.}
Table~\ref{tab:depth_object_size_analysis} shows that the lower performance of \textit{DiSR} is caused by samples with small target objects. For the smallest-object group, where the objects occupy only \textbf{0.236\%} of the image area, Qwen3-VL-8B-Instruct outperforms DiSR-8B-LoRA by 5.00\%, while the 3D-evidence construction error rate reaches 15.00\%. In contrast, for the largest-object group, where objects occupy 14.453\% of the image area, \textit{DiSR} achieves comparable accuracy to Qwen3-VL-8B-Instruct, with the 3D-evidence construction error rate reduced to only 1.67\%.
Figure~\ref{fig:depth_small_object_evidence_example} presents four representative failure cases involving small image-plane target objects. Across these examples, the queried objects occupy only a limited image region, making object-level depth and geometric estimation particularly challenging. All four examples are selected from samples that remain incorrect under GT Object Grounding but are recovered with GT 3D Evidence. This pattern indicates that the remaining errors arise from inaccurate object-level geometry rather than incorrect object references. These examples qualitatively support the size-conditioned trend reported in
Table~\ref{tab:depth_object_size_analysis}.

\section{Reproducibility Details}
\label{app:reproducibility}

\paragraph{Optimization.}
We use a QLoRA-style setup in both stages. The frozen backbone is
loaded with 4-bit NF4 quantization, double quantization, and bfloat16
compute, while only the LoRA parameters are optimized. LoRA is applied
to \texttt{q\_proj}, \texttt{k\_proj}, \texttt{v\_proj},
\texttt{o\_proj}, \texttt{gate\_proj}, \texttt{up\_proj}, and
\texttt{down\_proj}, with rank 128, scaling factor 256, dropout 0.05,
and no trainable bias. We use \texttt{paged\_adamw\_8bit} with
$\beta_1=0.9$, $\beta_2=0.999$, $\epsilon=10^{-8}$, zero weight
decay, and a maximum gradient norm of 1.0. The learning rate follows a
cosine schedule after a 3\% linear warmup. Training uses bfloat16,
gradient checkpointing, and a single GPU. The random seed is set to 42.

\paragraph{Perception Modules.}
All expert perception models remain frozen during training and
evaluation. The stack uses SAM~2.1 HQ with a Hiera-L backbone for object
masks, Metric3D~v2 with a ViT-Giant backbone for metric depth,
Perspective Fields with the ParamNet-360Cities checkpoint and
WildCamera with its all-data checkpoint for camera geometry, and
Orient Anything with a DINOv2-L backbone for object orientation. No
expert model is fine-tuned on the evaluation benchmarks.

\paragraph{Software and Compute.}
DiSR-8B is trained with Python 3.10.20, PyTorch 2.6.0+cu118,
Transformers 4.57.6, PEFT 0.19.1, Accelerate 1.14.0, and
bitsandbytes 0.49.2 on a single NVIDIA RTX 4080 SUPER under Linux
kernel 5.15.0-78-generic. Stages~1 and~2 require 19.50 and 39.25
GPU-hours, respectively, for a total of 58.75 GPU-hours.

\begin{table}[t]
\centering
\caption{
Training hyperparameters for the reported DiSR-8B model.
}
\label{tab:training_settings}
\small
\setlength{\tabcolsep}{5.5pt}
\renewcommand{\arraystretch}{1.05}
\begin{tabular}{lc}
\toprule
Setting & DiSR-8B \\
\midrule
Stage-1 training instances & 112,157 \\
Stage-2 training instances & 216,000 \\
Epochs per stage & 1 \\
Maximum sequence length & 1,024 \\
Per-device batch size & 1 \\
Gradient accumulation & 16 \\
Effective batch size & 16 \\
Stage-1 learning rate & $2\times10^{-5}$ \\
Stage-2 learning rate & $2\times10^{-5}$ \\
Stage-1 optimizer steps & 7,010 \\
Stage-2 optimizer steps & 13,500 \\
Warmup ratio & 0.03 \\
Stage-1 warmup steps & 211 \\
Stage-2 warmup steps & 405 \\
Optimizer & \texttt{paged\_adamw\_8bit} \\
Learning-rate scheduler & cosine \\
Weight decay & 0 \\
Maximum gradient norm & 1.0 \\
LoRA rank / scaling & 128 / 256 \\
LoRA dropout & 0.05 \\
Random seed & 42 \\
\bottomrule
\end{tabular}
\end{table}

\section{Error Analysis on SPAR-Bench}
\label{app:task_focused_training}

The main paper shows that the original training distribution provides limited supervision for relational-selection tasks in SPAR-Bench. To investigate whether targeted supervision can alleviate this limitation, we conduct an additional post-freeze diagnostic experiment. Starting from the frozen Epoch-2 \textit{DiSR-8B-LoRA} model, we construct a 2,816-example adaptation set, with half of the examples covering two previously unseen SPAR-Bench task types and the other half replaying the original spatial reasoning data to mitigate forgetting. We train a new LoRA adapter for 10 epochs and denote the resulting model as \textit{DiSR-8B-LoRA*}. We evaluate this model using the same full-chain protocol and report the results separately.

\paragraph{Results Discussion.}
As shown in Table~\ref{tab:spar_single_view}, \textit{DiSR-8B-LoRA*} improves over \textit{DiSR-8B-LoRA} by \textbf{11.23\%} on SPAR-Bench, demonstrating that task-specific supervision effectively addresses the identified limitation. Meanwhile, Table~\ref{tab:taskadapt_3dsrbench_fine} presents a fine-grained comparison on 3DSRBench, where \textit{DiSR-8B-LoRA*} maintains comparable overall performance with \textit{DiSR-8B-LoRA}, indicating that the additional task-focused adaptation improves SPAR-Bench performance without sacrificing general spatial reasoning ability.

\begin{table*}[t]
\centering
\caption{
Fine-grained relation accuracy (\%) on 3DSRBench, including the
task-adapted model. The best and second-best results in each row are shown in
\textbf{bold} and \underline{underlined}, respectively.
}
\label{tab:taskadapt_3dsrbench_fine}
\small
\setlength{\tabcolsep}{3.8pt}
\renewcommand{\arraystretch}{1.03}
\resizebox{\textwidth}{!}{%
\begin{tabular}{llccccc}
\toprule
Category
& Relation
& SpatialReasoner
& HiSpatial-3B
& DiSR-Qwen2.5
& DiSR-8B-LoRA
& DiSR-8B-LoRA* \\
\midrule

Height
& Higher
& 51.70 & \textbf{70.29} & 60.43 & \underline{62.75} & 62.61 \\

\midrule
Location
& Above
& \textbf{70.70} & 47.71 & 69.36 & \underline{70.38} & 63.29 \\

& Closer to camera
& 80.10 & \textbf{86.29} & 76.70 & \underline{80.68} & \underline{80.68} \\

& Next to
& \textbf{75.50} & 62.86 & 66.96 & 62.24 & 61.36 \\

\midrule
Orientation
& In front of
& 63.70 & \textbf{77.71} & 71.22 & \underline{75.29} & 73.26 \\

& On the left
& 64.20 & \textbf{85.14} & 70.20 & 77.08 & \underline{78.51} \\

& Viewpoint
& 36.40 & \textbf{67.43} & 53.35 & \underline{55.69} & 51.60 \\

\midrule
Multi-Object
& Closer to
& 54.90 & 73.14 & 75.14 & \underline{77.14} & \textbf{78.57} \\

& Facing
& \underline{68.20} & 52.57 & 67.34 & \textbf{70.81} & 67.63 \\

& Viewpoint toward object
& 34.40 & 32.57 & 46.94 & \textbf{50.73} & \underline{49.27} \\

& Parallel
& 50.10 & 47.43 & \textbf{62.54} & \underline{57.82} & 54.28 \\

& Same direction
& 55.50 & 50.29 & 56.10 & \textbf{59.59} & \underline{59.30} \\

\bottomrule
\end{tabular}%
}
\end{table*}

\clearpage

\end{document}